\documentclass{article}
\usepackage{spconf,amsmath,graphicx}
\usepackage{amsfonts,amssymb}
\usepackage{algorithm,algpseudocode}
\usepackage{booktabs}
\usepackage{amsmath}  
\usepackage{multirow}

\title{Efficient Universal Shuffle Attack for Visual Object Tracking}
\name{Siao Liu$^1$, Zhaoyu Chen$^1$, Wei Li$^1$, Jiwei Zhu$^1$, Jiafeng Wang$^2$, Wenqiang Zhang$^{1,2,*}$, Zhongxue Gan$^{1,3,*}$\thanks{This work was supported in part by Ji Hua Laboratory (project ID X190021TB190), in part by Shanghai Municipal Science and Technology Major Project (No.2021SHZDZX0103), in part by Science and Technology Commission of Shanghai Municipality (No.19511132000), in part by the Shanghai Engineering Research Center of AI and Robotics, and the Engineering Research Center of AI and Robotics, Ministry of Education, China.}}
\address{$^1$Academy for Engineering and Technology, Fudan University, Shanghai, China \\
	$^2$School of Computer Science, Fudan University, Shanghai, China\\
	$^3$The Department of Engineering Research Center for Intelligent Robotics, Ji Hua Laboratory, Foshan, China}
\begin{document}
%\ninept
%
\maketitle

\begin{abstract}
% Visual object tracking (VOT) has achieved great progress by deep neural networks (DNNs) and is widely used in downstream vision applications.
% Unfortunately, recent work shows that visual object tracking is vulnerable to adversarial examples, in which adversaries introduce delicate imperceptible perturbations to the input and fool DNNs to make incorrect predictions. 
% Unfortunately, there is many work demonstrating the vulnerability of VOT to adversarial attacks. 
% Currently, deep neural networks are shown to be vulnerable to adversarial attacks.  using imperceptible perturbation to fool the 
Recently, adversarial attacks have been applied in visual object tracking to deceive deep trackers by injecting imperceptible perturbations into video frames. However, previous work only generates the video-specific perturbations, which restricts its application scenarios. In addition, existing attacks are difficult to implement in reality due to the real-time of tracking and the re-initialization mechanism. To address these issues, we propose an offline universal adversarial attack called Efficient Universal Shuffle Attack. It takes only one perturbation to cause the tracker malfunction on all videos. To improve the computational efficiency and attack performance, we propose a greedy gradient strategy and a triple loss to efficiently capture and attack model-specific feature representations through the gradients. Experimental results show that EUSA can significantly reduce the performance of state-of-the-art trackers on OTB2015 and VOT2018.
\end{abstract}
\section{Introduction}
\label{sec:intro}
% denifition of vot and its success
Visual object tracking (VOT) is one of the fundamental vision tasks, which tracks a given object in each frame. It is the key building block in downstream vision applications, such as video surveillance and automatic driving. Thanks to the deep neural networks (DNNs), visual object tracking has achieved great progress. At present, the most representative algorithm is based on Siamese networks, which provides state-of-the-art performance and real-time inference, such as SiamFC~\cite{Siamfc}, SiamRPN~\cite{siamrpn}, SiamRPN++~\cite{siamrpnpp} and SiamMask~\cite{siammask}.

% introduce adversarial examples and fault in previous work
DNNs have been shown to be susceptible to adversarial examples. When an adversary introduces a delicate imperceptible perturbation to inputs, it would misguide the networks to produce incorrect results~\cite{L_BFGS,rpattck}. Previous adversarial attacks on visual object tracking~\cite{OA,CSA,hijack,spark,fan} have revealed the vulnerability of VOT. Online attacks, such as Cooling-Shrinking Attack (CSA)~\cite{CSA} and Hijacking~\cite{hijack}, generate the frame-specific perturbations online as the search region changes. Another attack is the offline attack, such as One-shot Attack (OA)~\cite{OA}. The perturbation is calculated in advance and directly added to the videos. It is worth noting that online attacks are difficult to implement in reality, as the object tracking happens in real time and has little time to generate specific perturbations for each frame.

Therefore, we focus on the offline attack for visual object tracking in this paper. 
% OA~\cite{OA} first proposes the offline attack.
The offline attack is first presented in One-shot Attack~\cite{OA} and it pre-generates perturbations on template images to fool the trackers. However, it ignores the re-initialization mechanism in tracking. When the tracker loses objects, it will reset a new template image as the target object, which leads to the One-shot Attack ineffective. The perturbation generated by One-shot Attack is target-specific, so it has low generalization and can hardly take effect on other videos, which restrains its application scenarios.

\begin{figure}[tb]
    \centering
    \includegraphics[scale=0.18]{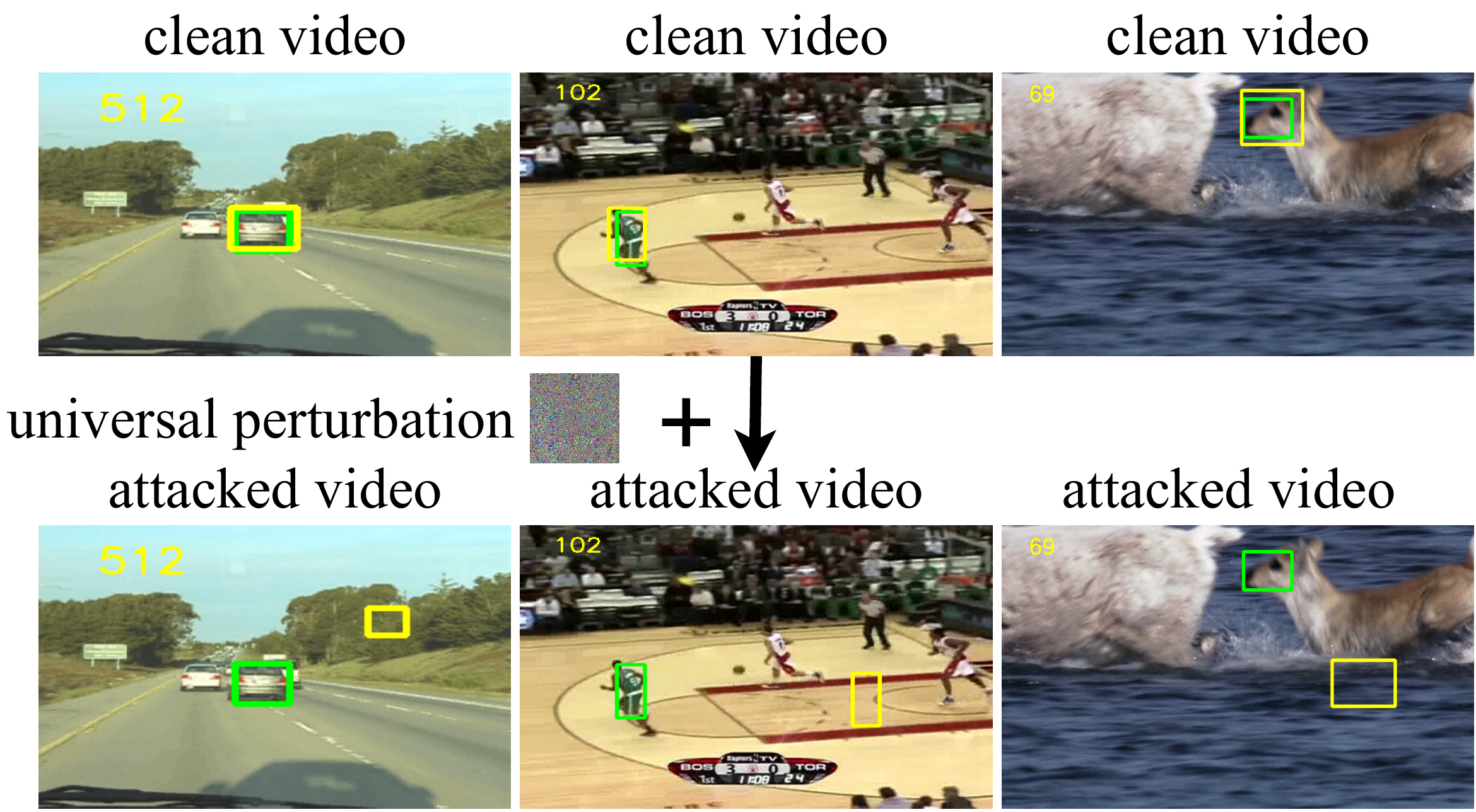}
    \caption{Visualization of the EUSA against visual object tracking. The green boxes represent the ground truth and the yellow boxes are the tracking result of the tracker.}
    \label{vis}
\end{figure}

% our work
To solve these issues, we propose a universal adversarial attack called Efficient Universal Shuffle Attack (EUSA). Specifically, we sample the dataset and only use a small part of videos to efficiently generate a video-agnostic and template-insensitive perturbation. To improve the attack performance, we design a triple loss for Siamese trackers from the perspectives of feature, confidence and shape. Moreover, we propose a greedy-gradient strategy to improve the sampling process. Greedy-gradient strategy captures model-specific feature representations through the gradients and selects the vulnerable videos. Fig.~\ref{vis} shows the effect of our proposed EUSA on OTB2015~\cite{OTB100}. Experiments show that EUSA effectively reduces the performance of state-of-the-art Siamese trackers on OTB2015~\cite{OTB100} and VOT2018~\cite{VOT2018}. Our major contributions can be summarized as follows:
\begin{itemize}
    \item We propose a novel universal adversarial perturbation generation method called Efficient Universal Shuffle Attack (EUSA). It takes only one perturbation to cause the tracker malfunction on all videos. 
    \item To improve the computational efficiency and attack performance, we propose a greedy-gradient strategy and a triple loss to effectively capture and attack model-specific feature representations through the gradients.
    \item Experimental results show that EUSA can efficiently and significantly reduce the performance of state-of-the-art Siamese trackers on various benchmarks.
\end{itemize}

\section{Related Work}
\subsection{Visual Object Tracking}
Recent work~\cite{Siamfc,siamrpn, siamrpnpp, siammask} based on Siamese networks has achieved excellent performance and real-time inference. SiamFC~\cite{Siamfc} first constructs a fully convolutional Siamese network to train a tracker. SiamRPN~\cite{siamrpn} introduces Region Proposal Network into tracking. Then SiamRPN++~\cite{siamrpnpp} proposes a multi-layer aggregation module and a depthwise correlation layer to achieve promising results with deeper networks. Moreover, SiamMask~\cite{siammask} enables Siamese trackers to conduct class-agnostic segmentation masks of the target object and improve the tracking performance. 

 \subsection{Adversarial Attacks for Visual Object Tracking}
Adversarial attacks on VOT~\cite{OA, CSA, hijack,spark,fan} can be divided into online and offline attacks, depending on whether perturbations are generated in real-time tracking or not. 
Both CSA~\cite{CSA} and Hijacking~\cite{hijack} are online attacks. The former uses U-Net~\cite{unet} to generate perturbations, while the latter directly uses gradient iterative calculation. However, these perturbations can not be generated in real-time tracking.
One-shot Attack (OA), the only offline attack known by us, ignores the re-initialization mechanism in tracking and only generates target-specific perturbations, which limits its attack performance.
To solve these issues, we propose the Efficient Universal Shuffle Attack to achieve offline universal attack.

\begin{figure}[tb]
    \centering
    \includegraphics[scale=0.17]{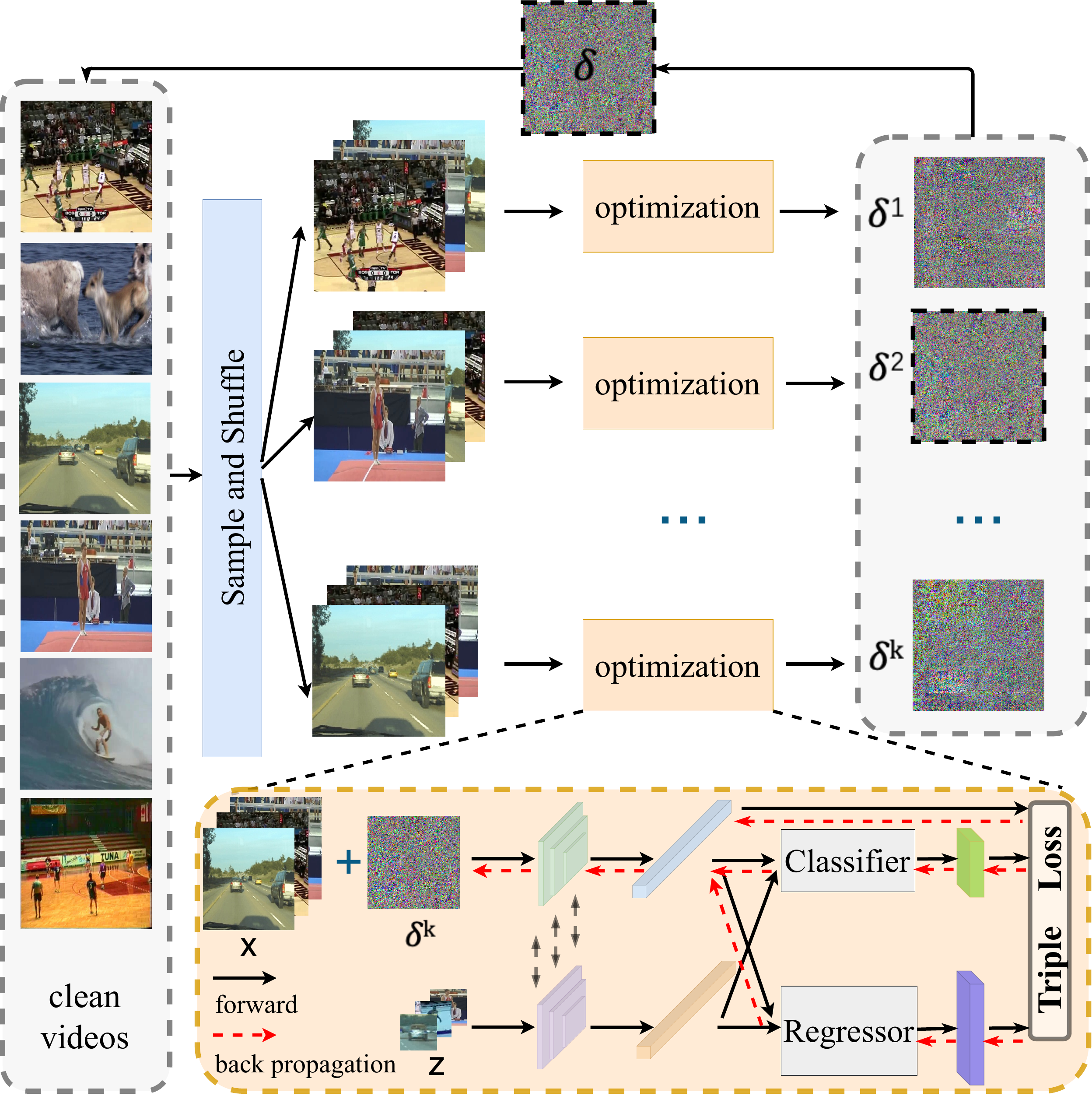}
    \caption{The overview of Efficient Universal Shuffle Attack.}
    \label{simple}
\end{figure}

\section{Methodology}

In this section, we first give the problem definition of universal attack on Siamese trackers and then we elaborate our Efficient Universal Shuffle Attack in detail.

\subsection{Problem Definition}
Given an unknown target template $z$, Siamese trackers need to predict the location and shape of the target in the subsequent frames. We decompose the Siamese tracker into three parts: feature encoder $\mathcal {F}$,  classifier $\mathcal {C}$ and bounding box regressor $\mathcal {R}$. The Siamese tracker shares the feature encoder $\mathcal{F}$ and conducts the similarity map between the template image $z$ and the search region $x$. Using the similarity map, classifier $\mathcal{C}$ obtains the confidence of candidate boxes to categorize the foreground and background.
Finally the bounding box regressor $\mathcal {R}$ conducts the location offset $R_{loc}$ and the shape offset $R_{shape}$ to adjust the location and shape of candidate boxes.

Considering the re-initialization mechanism, we aim to generate a universal perturbation $\delta$ on search regions to mislead a tracker to lose targets in all videos. Specifically, we describe the universal adversarial attack as follows:
\begin{equation}
   \mathop{\max} \sum_{x \in \mathcal{X}}\mathcal{L}(x,x+\delta), \quad s.t.\quad ||\delta||_{\infty} \leq \epsilon,
\end{equation}
where $\mathcal{X}$ are search regions from various videos and we limit the adversarial perturbation into the range $[-\epsilon,\epsilon]$, ensuring the perturbation is imperceptible to human eyes. For simplicity, we define $x^*$ as $x+\delta$, and $R_{loc}^*$ and $R_{shape}^*$ are the corresponding location offset and shape offset respectively. 

% For \textit{naturalness}, .
% For \textit{effectiveness}, we propose EUSA based on triple loss to attack all components of trackers and force the prediction target along the given direction, which is discribed in below. 

\subsection{Triple Loss}
To attack model-specific feature representations and improve attack performance, we design a triple loss from different perspectives, which combines feature-deflect loss $\mathcal{L}_{f}$, confidence loss $\mathcal{L}_{c}$ and drift loss $\mathcal{L}_{d}$. Supposing that $\lambda_1$ and $\lambda_2$ are the balanced weights, the triple loss $\mathcal{L}$ can be expressed as:
\begin{equation}
    \mathcal{L} = \mathcal{L}_{f} + \lambda_1\mathcal{L}_{c} + \lambda_2\mathcal{L}_{d} .
    \label{combine}
\end{equation}
 
 To attack the feature encoder and confuse similarity maps, we distort the embedding of the search region $\mathcal{F}(x)$ in feature space. We adopt the cosine similarity to measure the deflection of $\mathcal{F}_{i=1:C}(x)$, where $C$ is the channel of feature maps and set a margin $\mathrm{m_f}$ to control the deflection. Feature-deflect loss $\mathcal{L}_f$ can be written as:
\begin{equation}
    \mathcal{L}_{f}(x,x^*) = -\sum_{i=1:C} \max(\mathrm{m_f},\cos(\mathcal{F}_i(x),\mathcal{F}_i(x^*))).
\end{equation}

 Trackers always use Gaussian windows to constrain prediction results so they do not move too far within two adjacent frames, thereby providing some protection against attacks by improving background confidence. To address this issue, we suppress the confidence of all $N$ candidates and the confidence loss $\mathcal{L}_{c}$ is defined as follows:
\begin{equation}
    \mathcal{L}_{c}(z,x^*) = - \sum_{j=1:N} C_{j}(\mathcal{F}(z),\mathcal{F}(x^*)).
\end{equation}.

As the attack on the confidence-wise and feature-wise are undirected, we propose a drift loss $\mathcal{L}_{d}$  to force the predict bounding box along the given direction and shape overtimes. Specifically, we shrink the scale factors $R^{*}_{shape}$ to 0 and make the location factors $R^{*}_{loc}$ closer to the given direction $\vec{d}$, which is measured by Euclidean distance. In addition, we set constant $\alpha = 0.6$ to balance the shrinking attack and offsetting attack. Drift loss $\mathcal{L}_d$ can be expressed as:
\begin{equation}
    \mathcal{L}_{d}(z,x^*) = -\alpha \cdot ||R^{*}_{scale}||_2-||<R^{*}_{loc},\vec{d}>||_2.
    \label{ld}
\end{equation}

\subsection{Efficient Universal Shuffle Attack}

UAP~\cite{UAP} introduces high-dimensional decision boundaries to explain the existence of universal perturbation, which can be regraded as model-specific feature representations. We use data sampling to accelerate the capture of model-specific features representations. Considering that the magnitude of the gradient can reflect the videos' sensitivity to adversarial examples, we propose a greedy-gradient strategy as sampling approach. For each video $\mathbf{v}$, we could obtain the target template $z$ and the first search region $x_1$ through the initial frame. The required gradients can be computed efficiently by back propagation via Eq.~\ref{combine}. Then, we choose the videos with larger absolute value of gradient to construct training set $\mathbf{X} \in \mathcal{X}$ . The size of the training set depends on the sampling rate $r$.

% As we can easily access the gradient of  $\delta^{\tau}$, we adopt the Project Gradient Descent~\cite{PGD} to optimize the perturbation $\delta^{\tau}$:
% \begin{equation}
%     \delta^{\tau} =\mathrm{clip}\left( \delta^{\tau}+{\alpha}\cdot\mathbf{sign}(\nabla_{ \delta^{\tau}}L^{\tau}), -\epsilon, \epsilon \right)
% \end{equation}
% where $\alpha$ is the step size and the clip operation satisfies the constraint $||\delta^{\tau}||_{\infty}\leq \epsilon$ in each step. 

Fig.~\ref{simple} shows the pipeline of EUSA and the attack procedure in Algorithm~\ref{algo_EUSA}.
First, we sample the video dataset with greedy-gradient strategy and collect the victim video set $\mathbf{X}$. To search a better universal adversarial perturbation, we randomly shuffle the video set $k$ times to obtain the training set $\{\mathbf{X}^{1:k}\}$ and initialize $k$ perturbations$\{\delta^{\tau}\}_{\tau=1:k}$ as candidate perturbations. Since all candidate perturbations are optimized in the same way, we only discuss one perturbation $\delta^{\tau}$ below. Second, we randomly select one frame from each video in the shuffled training set $\mathbf{X}^{\tau}$ for iterative optimization. We follow the Project Gradient Descent~\cite{PGD} to optimize $\delta^{\tau}$ with Eq.~\ref{pgd}. The loss is calculated according to Eq.~\ref{combine}. 
As we can easily access the gradient of  $\delta^{\tau}$, we 
% adopt the Project Gradient Descent~\cite{PGD} to 
optimize the perturbation $\delta^{\tau}$ as follows:
\begin{equation}
    \delta^{\tau} =\mathrm{clip}\left( \delta^{\tau}+{\alpha}\cdot\mathbf{sign}(\nabla_{ \delta^{\tau}}L), -\epsilon, \epsilon \right)
\label{pgd}
\end{equation}
where $\alpha$ is the step size and the clip operation satisfies the constraint $||\delta^{\tau}||_{\infty}\leq \epsilon$ in each step. Then we sort all candidate perturbations by the loss $L^{\tau}$ which is the sum of triple loss $L^{\mathbf{v}}$ from selected videos. Finally, we choose the perturbation with the maximal loss $L^{\tau}$ as the final perturbation.

% \indent
% \textbf{greedy-Gradient Strategy}. 
% Dataset sampling guarantees our attack can achieve excellent performance with few videos. 
% Considering the magnitude of the gradient can reflect the sensitivity to adversarial examples, we propose a greedy-gradient strategy to select sensitive videos as victim data.

% \indent
% \textbf{Perturbation Optimization}.

\begin{algorithm}[tb]
\caption{Efficient Universal Shuffle Attack (EUSA)}
\label{algo_EUSA}
\begin{algorithmic}[1]
\Require 
 dataset $\mathcal{X}$, sampling rate $r$, victim tracker $\mathcal{T}$, number of candidate perturbations $k$
\Ensure
video-agnostic perturbation $\delta$
\State $\mathbf{X} \gets$ sampling ($\mathcal{X},r$) with \textbf{greedy-gradient strategy}
\State ${\mathbf{X}^{1:k}} \gets$  shuffle training set $\mathbf{X}$ for $k$ times
\State \textbf{Initialize} the $L^{best} \gets 0, \tau \gets 0$
\While {number of candidates $\tau++ < k$}
\State  \textbf{Initialize} $\tau$th perturbation $\delta^{\tau}$ with 0
\State \textbf{Initialize} the $\tau$th Loss $L^{\tau} \gets 0$
\For{$\mathbf{v}$ in $\mathbf{X}^{\tau}$}
\State \textbf{Initialize} attacked tracker $\mathcal{T}$ using template $z$
\State Randomly select a frame $I$
\State Obtaining the search region $s$ according $I$ and $\mathcal{T}$
\State Calculate the \textbf{triple loss} $L^{\mathbf{v}}$ using Eq.~\ref{combine}
\State $
    \delta^{\tau} \gets\mathrm{clip}\left( \delta^{\tau}+{\alpha}\cdot\mathbf{sign}(\nabla_{ \delta^{\tau}}L^{\mathbf{v}}), -\epsilon, \epsilon \right)
$
\State $L^{\tau} \gets L^{\tau} + L^{\mathbf{v}}$
\EndFor
\EndWhile
\State $\delta \gets \delta^{w} \quad \mathrm{s.t.}$ $w = \mathop{\arg\max} \{L^1,L^2,\dots,L^k\}$
\State \Return $\delta$
\end{algorithmic}
\end{algorithm}

\section{experiments}
We evaluate EUSA against state-of-the-art Siamese trackers on OTB2015 and VOT2018. Ablation studies show the effectiveness of greedy-gradient strategy and triple loss.

\subsection{Implementation Details}
We measure the performance of EUSA on two standard benchmarks, OTB2015~\cite{OTB100} and VOT2018~\cite{VOT2018}. OTB2015~\cite{OTB100} contains 100 videos and evaluates the trackers with precision and success rate. VOT2018~\cite{VOT2018} includes 60 videos and ranks the performance of trackers with the 
expected average overlap (EAO) rule.  Notably, the tackers would 
re-initialize the template image once it loses the target in VOT2018. The victim trackers include SiamRPN~\cite{siamrpn}, SiamRPN++~\cite{siamrpnpp} and SiamMask~\cite{siammask}. SiamRPN++(R) represents that the SiamRPN++ applies ResNet-50~\cite{resnet} as the backbone and the SiamRPN++(M) uses MobileNet-v2~\cite{mobilenetv2} to extract features.
We implement experiments with RTX 1080Ti. The number of candidate perturbations $k$ is set to 50. For each iterative step, we set step size $\alpha=0.9$ and the maximum pixel of perturbation $\epsilon=16$. To balance the components of triple loss, we set the hyper-parameters $\lambda_1 = 0.9, \lambda_2 = 0.7$.
In all experiments, we report the average result over 5 times. 

\begin{table}[t]
\centering
\caption{Attack performance on OTB100.}
\label{resultOTB}
\scalebox{0.87}{
\begin{tabular}{@{}l|cccccc@{}}
\toprule
\multicolumn{1}{c|}{\multirow{2}{*}{Tracker}} & \multicolumn{3}{c|}{Precision(\%) $\uparrow$}                         & \multicolumn{3}{c}{Success Rate(\%) $\uparrow$ } \\ \cmidrule(l){2-7} 
\multicolumn{1}{c|}{}                         & Org  & OA   & \multicolumn{1}{c|}{EUSA}                                        & Org  & OA   & EUSA                   \\ \midrule
\multicolumn{1}{l|}{SiamRPN}                 & 87.6 & 27.8 & \multicolumn{1}{c|}{\textbf{26.7}} & 66.8 & 20.4 & \textbf{20.2} \\
\multicolumn{1}{l|}{SiamRPN++(R)}            & 90.5 & 35.7 & \multicolumn{1}{c|}{\textbf{32.7}} & 69.6 & 26.2 & \textbf{23.6} \\
\multicolumn{1}{l|}{SiamRPN++(M)}            & 86.4 & 35.3 & \multicolumn{1}{c|}{\textbf{25.9}} & 65.8 & 26.1 & \textbf{18.3} \\
\multicolumn{1}{l|}{SiamMask}                & 83.9 & 65.0 & \multicolumn{1}{c|}{\textbf{34.9}} & 64.7 & 48.1 & \textbf{22.5} \\ \bottomrule
\end{tabular}
}
\caption{Attack performance on VOT2018.}
\label{resultVOT2018}
\scalebox{0.6}{
\begin{tabular}{@{}l|cccccclll@{}}
\toprule
\multicolumn{1}{c|}{\multirow{2}{*}{Tracker}} & \multicolumn{3}{c|}{Accuracy(\%) $\uparrow$}                          & \multicolumn{3}{c|}{Robustness $\downarrow$}                               & \multicolumn{3}{c}{EAO $\uparrow$}                 \\ \cmidrule(l){2-10} 
\multicolumn{1}{c|}{}                         & Org  & OA   & \multicolumn{1}{c|}{EUSA}                                        & Org   & OA    & \multicolumn{1}{c|}{EUSA}                                          & Org   & OA    & EUSA                    \\ \midrule
\multicolumn{1}{l|}{SiamRPN}                 & 57.7 & 46.7 & \multicolumn{1}{c|}{\textbf{44.0}} & 0.309 & 1.733 & \multicolumn{1}{c|}{\textbf{2.241}} & 0.338 & 0.082 & \textbf{0.055} \\
\multicolumn{1}{l|}{SiamRPN++(R)}            & 60.2 & 51.9 & \multicolumn{1}{c|}{\textbf{46.1}} & 0.243 & 1.157 & \multicolumn{1}{c|}{\textbf{2.051}} & 0.413 & 0.115 & \textbf{0.072} \\
\multicolumn{1}{l|}{SiamRPN++(M)}            & 58.9 & 48.3 & \multicolumn{1}{c|}{\textbf{45.2}} & 0.234 & 1.344 & \multicolumn{1}{c|}{\textbf{2.622}} & 0.411 & 0.101 & \textbf{0.056} \\
\multicolumn{1}{l|}{SiamMask}                & 59.8 & 45.5 & \multicolumn{1}{c|}{\textbf{31.8}} & 0.248 & 0.674 & \multicolumn{1}{c|}{\textbf{2.632}} & 0.406 & 0.165 & \textbf{0.043} \\ \bottomrule
\end{tabular}
}

\caption{Ablation study of triple loss.}
\label{ablation}
\scalebox{0.84}{
\begin{tabular}{@{}c|ccccccc@{}}
\toprule
$\mathcal{L}_f$              &      & \checkmark &      &      & \checkmark &     \checkmark & \checkmark \\
$\mathcal{L}_c$               &      &      & \checkmark &      & \checkmark &  & \checkmark \\
$\mathcal{L}_d$             &      &      &      & \checkmark &      & \checkmark & \checkmark \\ \midrule
\multicolumn{1}{c|}{Precision(\%)}     & 90.5 & 59.4 & 54.1 & 54.6 &51.2 &      50.0& \textbf{32.7} \\
\multicolumn{1}{c|}{Success rate(\%)}  & 69.6 & 40.0 & 38.4 & 39.0 &37.1  &      36.2 & \textbf{23.6} \\ \bottomrule
\end{tabular}
}
\end{table}

\begin{figure}[t]
\begin{minipage}[b]{.48\linewidth}
  \centering
  \centerline{\includegraphics[scale=0.2]{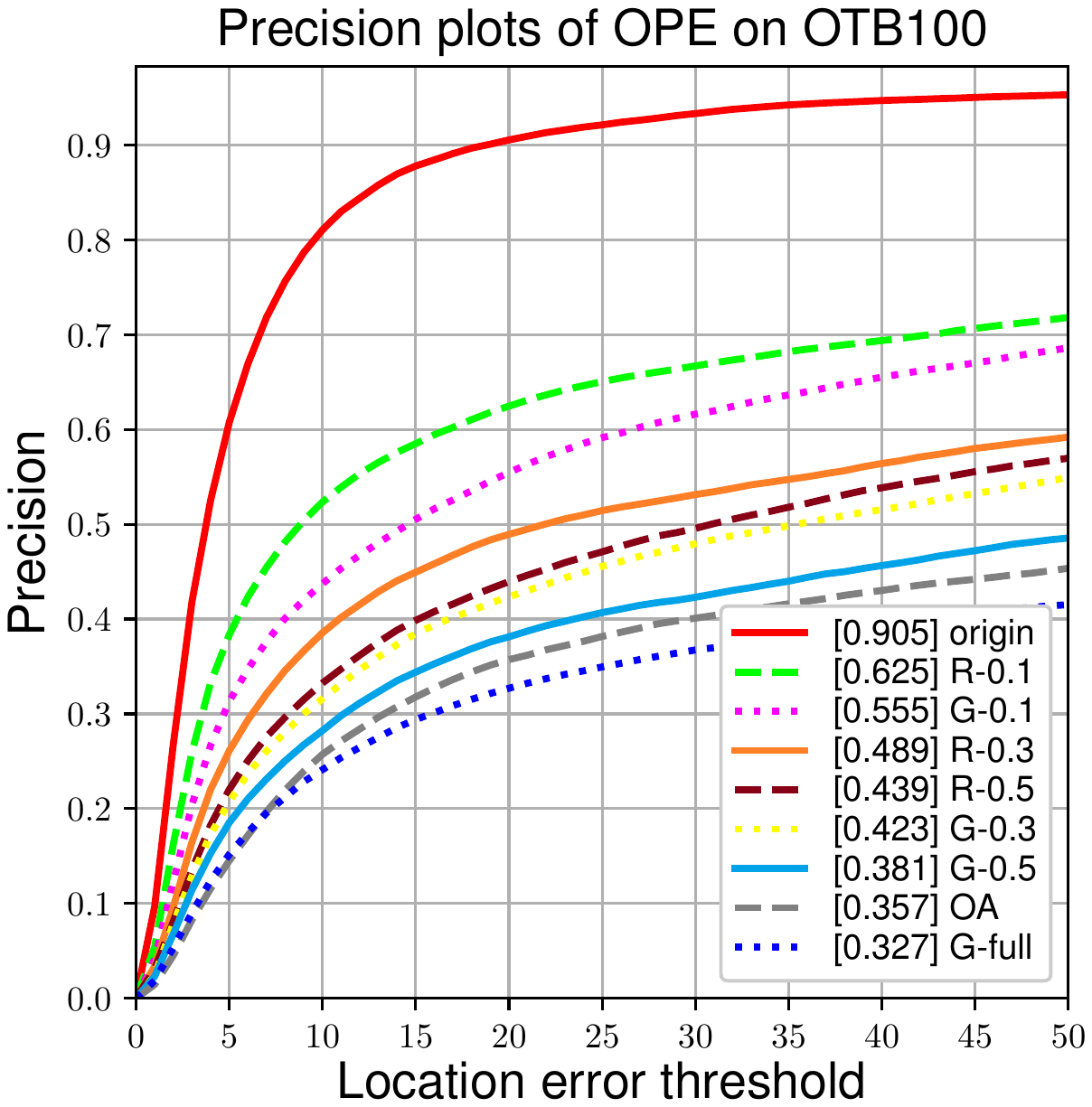}}
%  \vspace{1.5cm}
%   \centerline{(a) Results 3}\medskip
\end{minipage}
% \hfill
\begin{minipage}[b]{0.48\linewidth}
  \centering
  \centerline{\includegraphics[scale=0.2]{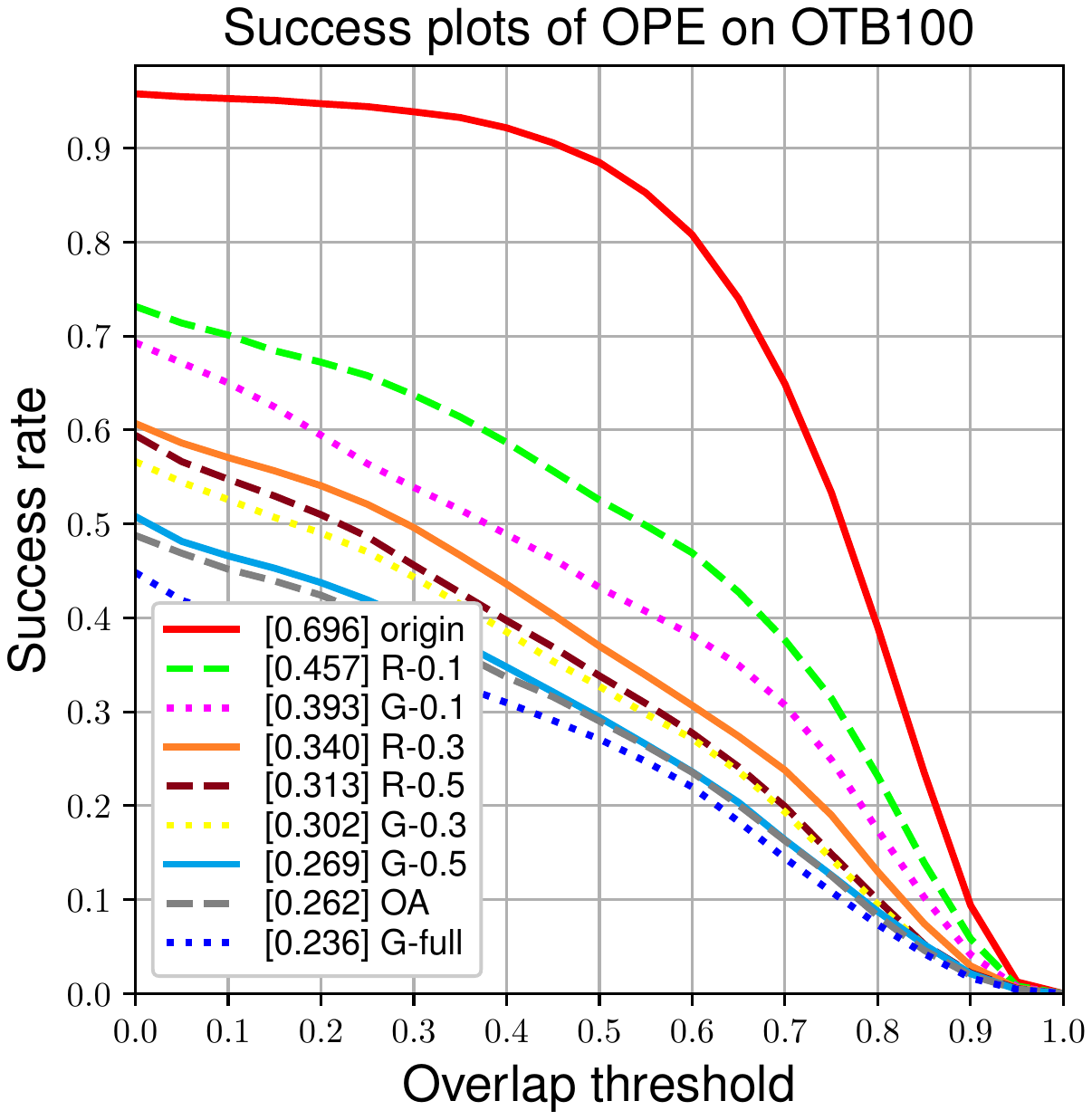}}
%  \vspace{1.5cm}
%   \centerline{(b) Result 4}\medskip
\end{minipage}
\caption{Quantitative comparisons between various sampling rate and different sampling strategy on OTB2015 dataset. The suffix "G" and "R" are greedy-gradient strategy and random sample respectively. The numbers are sampling rates.}
\label{PS}
\end{figure} 

\subsection{Attacks on OTB2015 and VOT2018}

\indent
\textbf{Results on OTB2015.} As shown in Table~\ref{resultOTB}, EUSA successfully drops the success rates of SiamRPN, SiamRPN++(R), SiamRPN++(M) and SiamMask to 20.2\%, 23.6\%, 18.3\% and 22.5\%. Besides, EUSA has the best performance on the SiamRPN++(M), which reduces the precision and success rate by 60.5\% and 47.5\% respectively. Compared with SiamRPN++(R), SiamRPN++(M) is more vulnerable to our universal adversarial perturbation, which can be attributed to the vulnerability of its backbones.

\noindent\textbf{Results on VOT2018.} Table~\ref{resultVOT2018} shows that EUSA performs much better than OA in VOT2018, which reveals that One-shot Attack is extremely template-sensitive. Once the trackers re-initialize the template, OA fails to attack the subsequent frames. However, slight distortion of template image almost has no impact on our attack, which can be contributed to the randomly shuffle. In addition, EUSA trains the perturbation by selecting the target template and search region from different frames, which makes the re-initialization mechanism ineffective against our attack.

\subsection{Ablation Study}
We conduct a series of experiments on OTB2015 to explore the effectiveness of each component in our attack. We use the SiamRPN++(R) as the victim tracker.

Fig.~\ref{PS} illustrates a quantitative analysis of the performance of EUSA with different sampling strategies and various sampling rates $r$, including 0.1, 0.3, 0.5, and 1. Our proposed EUSA with any sampling rate can significantly decline the performance of SiamRPN++(R), which indicates that our attack still performs well on the videos unseen in the training process. When the sampling rate is 0.1, the perturbation generated by only 10 frames can drop the tracker precision by 35\%. Compared with random sampling, our greedy-gradient strategy can significantly improve attack performance by at least 5.8\% on precision and 2.3\% on success rate. 

Moreover, we evaluate the performance of each component of the triple loss. The results are shown in Table~\ref{ablation}. We observe that only using the confidence loss would outperform the other two, which reduces the precision and success rate to 54.1\% and 38.4\% respectively. This shows that $\mathcal{L}_c$ successfully damages the Gaussian window and disables the Region Proposal Network. Besides, there is no performance degradation when simultaneously using any two components of triple loss which validates the effectiveness of our triple loss. 

\section{Conclusions}
In this work, we propose an offline universal attack, Efficient Universal Shuffle Attack (EUSA), which injects only one perturbation to cause the tracker malfunction on all videos. To further improve the efficiency and performance of EUSA, we design a greedy-gradient strategy and a triple loss to capture and attack the model-specific feature representations. Numerous experiments show that EUSA can significantly reduce the performance of Siamese trackers on various benchmarks.

% make more researchers pay attention to the adversarial attack on tracking algorithm. We Although good performance have been achieved in adversarial attacks, efficient attacks are still lacking and we hope that this work can bring more attention to lightweight attacks.
\clearpage
\bibliographystyle{IEEEbib}
\bibliography{strings,refs}
\end{document}